\documentclass[conference]{IEEEtran}

\usepackage{layout,color,times,helvet,multicol,multirow,lscape,rotating,fancyhdr,balance}
\usepackage{algorithm}
\usepackage{algorithmic}
\usepackage{graphicx}
\usepackage{setspace}
\usepackage{color}
\usepackage{amsmath}
\usepackage{lineno}
\usepackage{colortbl}
\usepackage[usenames,dvipsnames,table]{xcolor}

\usepackage{fancyhdr}
\begin{document}
\title{\vspace{-0.5cm}Self-Configuring and Evolving \\Fuzzy Image Thresholding}

\author{
\IEEEauthorblockN{A. Othman$^1$, H.R. Tizhoosh$^2$, F. Khalvati$^3$}
\IEEEauthorblockA{$^1$ Dept. of Information Systems, Computers \& Informatics, 
 Suez Canal University, Egypt, a.othman@ci.suez.edu.eg\\ $^2$ KIMIA Lab, 
University of Waterloo, Canada, tizhoosh@uwaterloo.ca\\ $^3$ Dept. of Medical Imaging, Sunnybrook Research Institute, University of Toronto, Canada, farzad.khalvati@sri.utoronto.ca}
}

\maketitle
\thispagestyle{fancy}
\vspace{-0.5cm}
\begin{abstract}
Every segmentation algorithm has parameters that need to be adjusted in order to achieve good results. Evolving fuzzy systems for adjustment of segmentation parameters have been proposed recently (Evolving fuzzy image segmentation -- EFIS \cite{EFIS}). However, similar to any other algorithm, EFIS too suffers from a few limitations when used in practice. As a major drawback, EFIS depends on detection of the object of interest for feature calculation, a task that is highly application-dependent. In this paper, a new version of EFIS is proposed to overcome these limitations. The new EFIS, called self-configuring EFIS (SC-EFIS), uses available training data to auto-configure the parameters that are fixed in EFIS. As well, the proposed SC-EFIS relies on a feature selection process that does not require the detection of a region of interest (ROI). 
\end{abstract}
\maketitle

\section{introduction}
Evolving fuzzy image segmentation has been recently introduced to solve the parameter setting problem (e.g., fine-tuning) of different segmentation techniques. EFIS has been designed with emphasis on acquiring and integrating user feedback into the fine-tuning process. As a result, EFIS is suitable for all those applications, such as medical image analysis, in which an experienced and knowledgeable user provides evaluative feedback of some sort with respect to the quality, i.e., accuracy, of the segmentation process. Image segmentation is the grouping of pixels to form meaningful clusters of pixels that constitute objects (e.g., organs, tumours), a task with various applications in medical image analysis including measurement, detection, and diagnosis. Image segmentation can be roughly categorized into two main classes of algorithms; non-parametric (e.g., atlas-based segmentation) and parametric (e.g., region growing) algorithms. The former is based on a model which usually does not require parameters whereas the latter is based on some parameters that must be adjusted in order to obtain reasonable segmentation results. Parameter-based segmentation algorithms always face the challenge of parameter adjustment; a parameter tuned for a particular set of images may perform poorly for a different image category. On the other hand, in a clinical setting such as a hospital, the final outcome of image segmentation algorithms usually need to be modified (i.e., manually edited) and approved by a an expert (e.g., radiologist, oncologist, pathologist). The clinical ramifications of not verifying the correctness of segments include missing a target (resulting in a less effective therapy) or increased toxicity if the target is over-segmented. The frequent expert intervention to correct the results, in fact, generates valuable feedback for a learning scheme to automatically adjust the segmentation parameters. EFIS is a segmentation scheme that evolves fuzzy rules to tune the parameters of a given segmentation algorithm by incorporating the user feedback which is provided to the system as corrected or manually created segmentation results called \emph{gold standard images}. EFIS represents a new understanding of how image segmentation should be designed in the context of observer-oriented applications. Naturally, EFIS needs to be further improved and extended in order to exploit the full potential of its underlaying evolving mechanism in relation to the user feedback. The original design of EFIS as presented in~\cite{EFIS} requires pre-configurations of a few steps which should be set for a given image set and the segmentation algorithm to which EFIS is integrated. This limits the efficiency of EFIS; either the algorithm should be pre-configured for each dataset and/or segmentation algorithm or it is possible that a fixed pre-configuration will adversely affect its performance. In this paper, we present an extended version of EFIS which we call self-configuring EFIS (short SC-EFIS) that has a higher level of automation. The new extension of EFIS proposed in this paper will enhance EFIS through removing these limitations by introducing self-configuration into different stages of EFIS.

This paper is organized as follows: In section~\ref{SummaryEFIS}, a brief review of the EFIS (evolving fuzzy image segmentation) will be provided. In section~\ref{SCEFISsection}, we present the proposed self-configuring EFIS (SC-EFIS). In section~\ref{expResults}, experiments are described and the results are presented and analyzed. Finally, section~\ref{CON} concludes the paper.

\section{A Brief Review of EFIS}
\label{SummaryEFIS}

The concept of Evolving Fuzzy Image Segmentation, EFIS, was proposed recently \cite{EFIS,EFIS1}. The problem that EFIS attempts to address is parameter adjustment in image segmentation. The basic idea of EFIS is to adjust the parameters of segmentation to increase the accuracy by using user feedback in form of corrected segments. To do so, EFIS extracts features from a region inside the image  and assigns them to the best parameters that are exhaustively detected in an offline stage or maintenance cycle. Clustering or other methods are then used to generate fuzzy rules, which are then continuously updated when new images are processed. EFIS needs to be trained for specific algorithms and image categories. In other words, in order to employ EFIS, the following components must be pre-designated:

\begin{itemize}
\item Parent algorithm: any segmentation algorithm with at least one parameter that affects its  accuracy 
\item Parameter(s) to be adjusted (e.g., thresholds, scales)
\item Images and corresponding gold standard images 
\item Procedure to find optimal parameters (e.g., brute force or trial-and-error via comparison with the gold standard images)
\end{itemize}

Once the above-mentioned components are made available, the following steps need to be specified in EFIS:

\begin{itemize}
\item ROI-detection algorithm: An algorithm that detects the region of interest (ROI) around the subject to be segmented by EFIS.
\item Procedure for feature extraction around available seed points: Methods like SIFT are used to generate seed points. But a certain number of expressive features should be calculated in the vicinity of each seed point to be fed to fuzzy inference system.
\item Rule pruning: Upon processing a new image, a new rule can be learned only if the features and corresponding output parameters had not been observed previously. In other words, by looking at the difference between an input  (features plus outputs) with all rules in the database, the information of a new image is added only if not captured by existing rules.
\item Label fusion: When EFIS is used with multiple algorithms at once, the segmentation results are fused using a fusion method namely STAPLE algorithm~\cite{Warfield2004}.
\end{itemize}

EFIS includes two main phases namely training and testing. In training phase, images with their gold standard results are fed to the algorithm where features are extracted from each image. The parent algorithm, e.g., thresholding \cite{nib,Huang1995,Kittler1986,Tizhoosh2005image, Rahnamayan2008,Shokri2003,Tizhoosh2009}, is applied to each image and the results are compared to the gold standard image. The algorithm's parameters are continuously changed until the best possible result is achieved. The parameter which yields the best result (i.e., the highest agreement with the gold standard image) along with the image feature extracted in the previous stage are stored. Once all training images are processed, the fuzzy rules are generated from the stored data using a clustering algorithm.

In testing phase, new images are first processed to extract features. Next, the image features are fed to the fuzzy inference system to approximate the parameters. The parent algorithm is then applied to the input image using the estimated parameter. EFIS can address both single-parametric and multi-parametric problems. EFIS was applied to three different thresholding algorithms  and significant improvements in terms of segmentation accuracy were achieved \cite{EFIS}.

\section{Self-Configuring EFIS (SC-EFIS)}
\label{SCEFISsection}
This section introduces a new version of EFIS, namely a self-configuring evolving fuzzy image segmentation (SC-EFIS) which represents a higher level of automation compared to the original EFIS scheme. The proposed SC-EFIS scheme consists of three phases;  self-configuration phase, training phase, and online or evolving phase. In the following, each of these phases is described in detail.

\subsection{Self-Configuring Phase}
\label{SCEFISPre}

In the self-configuring phase (Algorithm \ref{SCEFIS_Pre}), all available images are processed in order to determine two crucial factors: 1) the size of the feature area around each seed point, and 2) the final features to be used for the current image category.

\begin{figure}[htb]
\includegraphics[width=0.9\columnwidth]{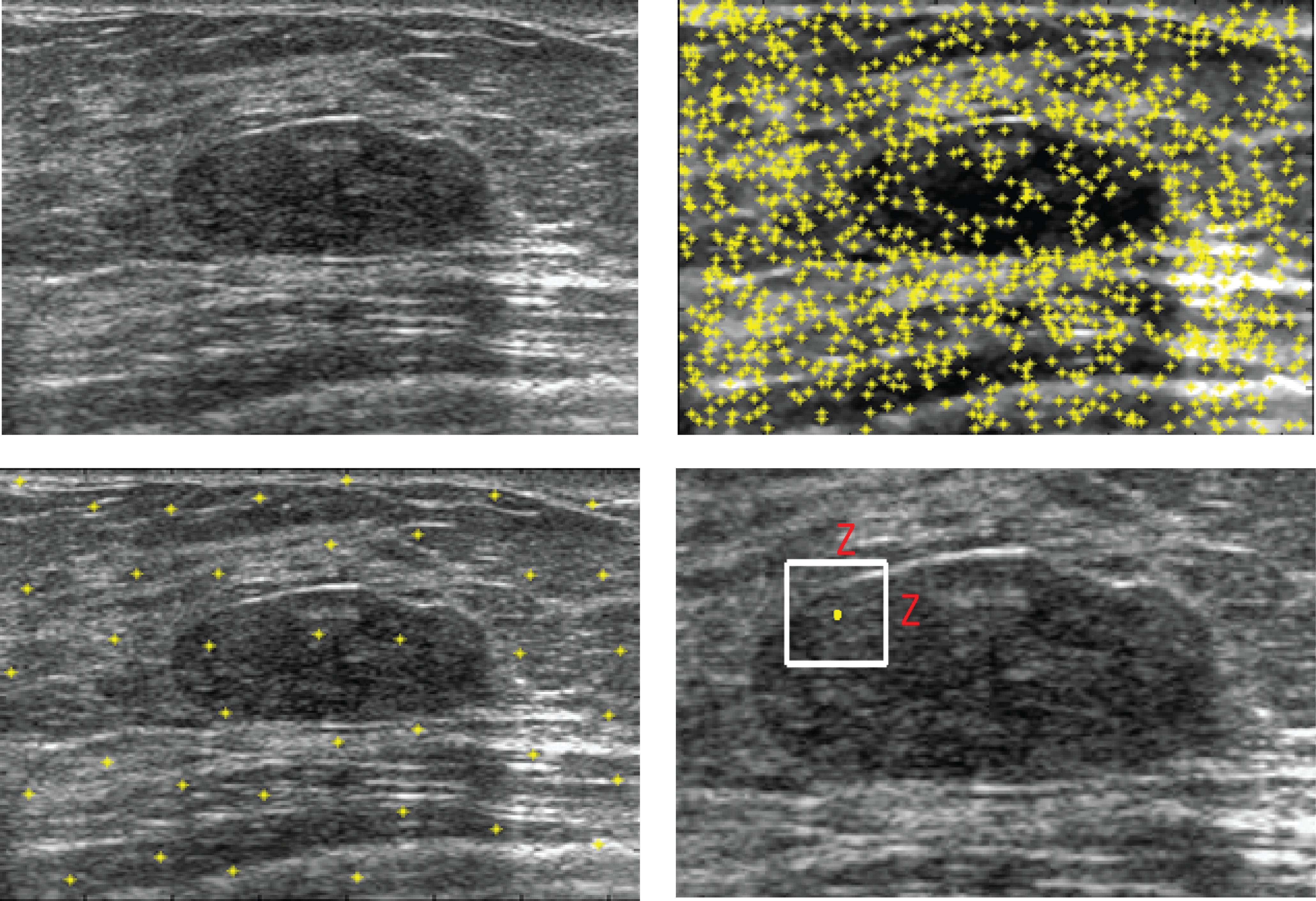}
\caption{Feature extraction process (from top left to bottom right): original image, seed points detected by SIFT, selected seed points via sorting the descriptor, calculating features around each selected seed point.}
\label{Extract}
\end{figure}

The $Z \times Z$ rectangle around each SIFT point to be used for feature calculation is determined based on different sizes of all available images (Algorithm \ref{SCEFIS_Pre}). Following this step, the set of features that should be used for the available images is selected from a large number of features which are calculated for each image from the vicinity of the SIFT points located in the entire image (since there is no longer an ROI) (Fig. \ref{Extract}). This process starts with the determination of the number of SIFT points $N_F$ that should be used in the current image (algorithm \ref{SCEFIS_Pre}). This step is identical to the procedure used in the EFIS training phase, as previously explained in section~\ref{SummaryEFIS}, with three exceptions: the SIFT points are detected across the entire image (as opposed to selecting SIFT points inside an ROI as a subset of the image), the final number $N_F$ of SIFT seed points is not fixed, and the points returned are separated from each other by $Z$ in each direction. For all $N_F$ seed points, features are extracted from a rectangle $R_C$ around each point, based on the discrete cosine transform ($D_C$) of $R_C$, the gradient magnitude ($G_M$) of $R_C$, the approximation coefficient matrix $A_C$ of $R_C$ (computed using the wavelet decomposition of $R_C$), and the SIFT descriptors $D_S$. The following set of features is extracted (Algorithm \ref{SCEFIS_Pre}):

\begin{enumerate}
\item The mean, median, standard deviation, co-variance, mode, range, minimum, and maximum of $R_C$, $D_{C_{R_C}}$, and $A_{C_{R_C}}$, and $G_{M_{R_C}}$ (32 features)

\item The mean, median, standard deviation, co-variance, range, minimum, maximum, and zero population of $D_S$ (eight features) with the minimum of $D_S$ changed to be the minimum number after zero

\item The contrast, correlation, energy, and homogeneity of the gray level co-occurrence matrices (computed in four directions $0\,^{\circ}$, $45\,^{\circ}$, $90\,^{\circ}$, and $135\,^{\circ}$) of $R_C$, $D_{C_{R_C}}$, and $A_{C_{R_C}}$, and $G_{M_{R_C}}$ (64 features)

\item The contrast, correlation, energy, and homogeneity of the gray level co-occurrence matrices (computed in only one directions of $0\,^{\circ}$) of $D_S$ (four features)

\item A feature matrix $F_1$ of size $N_F \times N_T$ generated for $I$ (in this case $N_T=108$)
\end{enumerate}

\begin{algorithm}[htb]
\caption{\textcolor{black}{Self-Configuration Phase}}
\begin{algorithmic}[1]
\label{SCEFIS_Pre}
\STATE Set the variables and initialize all matrices 
\STATE Read the available images $I_1, I_2, \cdots , I_{N_I}$.
\STATE Read the size of the images, namely all rows $R_1, R_2, \cdots , R_{N_I}$, and all columns $C_1, C_2, \cdots , C_{N_I}$.
\STATE Determine the size of the rectangle\\ Z = $0.1\times \textrm{max}(\textrm{median}_i(R_i),\textrm{median}_i(C_i))$.
\STATE Create the initial matrix  $F_1$ and the final matrix $F^*$.
\FOR{each image}
\STATE Determine $N_F$, the number of SIFT points, that should be used for image $I_i$.
\FOR{each SIFT point}
\STATE Extract features $f_1,f_2,\cdots,f_{N_T}$  from the $Z \times Z$ rectangle around each SIFT point.
\STATE Append the features as a new row to the initial matrix $F_1$, which becomes of size $N_F \times N_T$.
  \ENDFOR
\STATE Calculate $S_T$ different statistics from $F_1$ and assigned in $F_2$.
\STATE Append $F_2$ of the current image of size $S_T\times N_T$ to the feature matrix $F_3$ (the feature matrix $F_3$ becomes of size $L \times N_T$, $ L= S_T*N_I$)
\ENDFOR
\STATE Remove very similar features from $F_3$ (e.g., at least 99\% correlated). $F_4$ is a reduced matrix of $F_3$ of size $L \times N_{T_1}$, $N_{T_1} \leq N_T$.
\STATE Determine the number of features by discarding similar ones from $F_4$ (e.g., at least 90\% correlated). $F_C$ is a feature matrix generated from $F_4$ of size $L \times N_{T_2}$, $N_{T_2} \leq N_{T_1}$.
\STATE Use $k$ different unsupervised feature selection methods to generate $k$ different feature matrices in addition to $F_C$: $F_P$, $F_M$, $F_F$, $F_G$, and $F_L$. All of these matrices are of size $L \times N_{T_2}$.
\STATE Select any features found in at least half of the matrices to form $F_5$ of size $L \times N_{T_3}$, $N_{T_3} \leq N_{T_2}$.
\STATE Generate a final feature matrix $F^*$ from $F_5$ by removing similar features (e.g., at least 90\% correlated). $F^*$ is of size $L \times N_L$, $N_L \leq N_{T_3}$.
 \end{algorithmic}
 \end{algorithm}
 
The next step is to calculate $S_T$ different statistical measures from $F_1$ (e.g.,  $S_T=8$: mean, median, mode, standard deviation, co-variance, range, minimum, and maximum). The resulting matrix $F_2$ (size $S_T \times N_T$) is returned, in which each row represents a statistical measure (Algorithm \ref{SCEFIS_Pre}, \emph{CSF}). $F_2$ is then appended to the feature matrix $F_3$ (Algorithm \ref{SCEFIS_Pre}). After all images are processed, the feature matrix $F_3$ is formed from the features of all images, with each image being represented by $S_T$ rows. In the last step, the final set of features that should be used in the current image category are selected from $F_3$. This process starts with the removal of very similar features in $F_3$ based on the calculation of the correlations between all features. Hence, if two features are highly correlated, e.g. with a correlation coefficient of at least 99\%, then one is kept and the other is discarded. The output of this process is a matrix $F_4$  (Algorithm \ref{SCEFIS_Pre}).

\begin{figure}[tb]
\includegraphics[width=0.9\columnwidth]{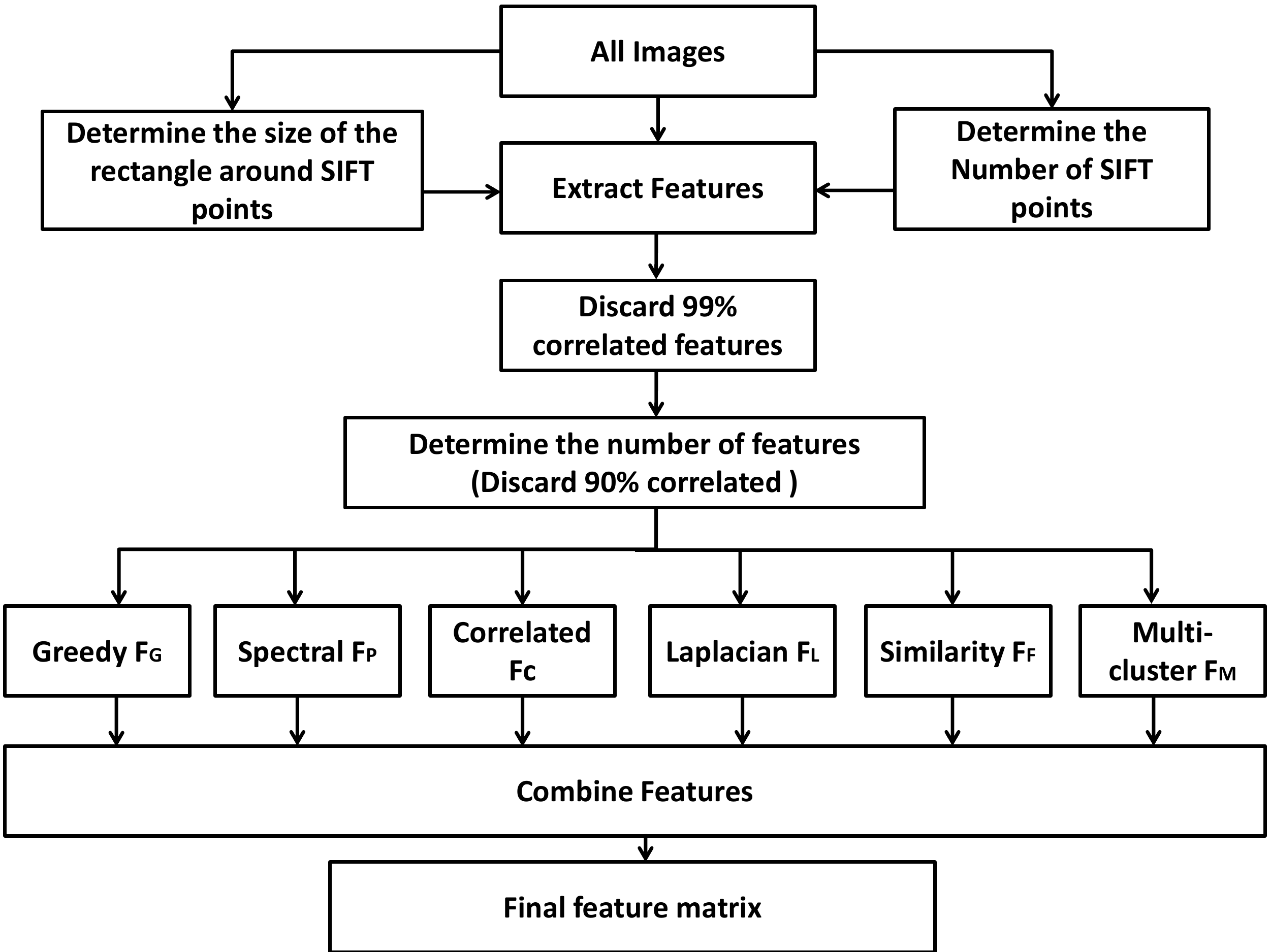}
\caption{The process of feature selection}
\label{allFeatures}
\end{figure}

Five methods, along with an additional correlation-based method, were combined to produce an ensemble of final relevant features that could be used for training (see Fig. \ref{allFeatures}): Mitra et al. \cite{mitra2002unsupervised}- $F_F$ (feature similarity), He et al. \cite{he2006laplacian}- $F_L$ (Laplacian score), Zhao et al. \cite{zhao2007spectral}- $F_P$ (spectral graph), Cai et al. \cite{cai2010unsupervised}- $F_M$ (multi-cluster),  Farahat et al. \cite{farahat2012efficient}- $F_G$ (greedy algorithm), and $F_C$ (correlation method).

 For any unsupervised feature selection technique, the number of features $N_{T_2}$ that should be returned must be established in advance. A correlation with a threshold of 90\% is used in order to determine the number of features that should be returned from $F_4$ (Algorithm \ref{SCEFIS_Pre}). Following this process, $F_C$ is the resulting feature matrix. In addition to $F_C$, five different unsupervised feature selection methods are also used for feature selection. The matrix $F_4$ and the variable $N_{T_2}$ are passed to the methods, and each method returns a different matrix with its selected features. The resulting matrices are $F_G$ \cite{farahat2012efficient}, $F_L$ \cite{he2006laplacian}, $F_F$ \cite{mitra2002unsupervised}, $F_P$ \cite{zhao2007spectral}, and $F_M$ \cite{cai2010unsupervised} (Algorithm \ref{SCEFIS_Pre}). For all features in the six matrices, any feature extracted by at least three of the six methods are selected and appended to a matrix $F_5$ (Algorithm \ref{SCEFIS_Pre}). The final matrix $F^*$ is generated based on the discarding of features from $F_5$ that are at least 90\% correlated (Algorithm \ref{SCEFIS_Pre}).
 
\subsection{Offline Phase}
In the offline phase, the best parameters for segmenting each image are calculated through an exhaustive search and then stored in matrix $T$ (Algorithm \ref{SCEFIS_TRAIN}). The process is performed as explained in \cite{EFIS}.

\subsection{Training Phase}
\label{SCEFISTrain}
In this phase, the features selected for the training images are used for the training of the fuzzy system. A set of images are randomly selected for training (Algorithm \ref{SCEFIS_TRAIN}). A matrix $M$ is created and filled with the rows from $F^*$ that belong to the training images (Algorithm \ref{SCEFIS_TRAIN}). A matrix $O$ is created and filled with the rows from $T$ that belong to the training images (Algorithm \ref{SCEFIS_TRAIN}). A pruning step is performed starting from the second training image in order to ensure that $M$ and $O$ do not contain similar rows (Algorithm \ref{SCEFIS_TRAIN}). The pruned matrices $M$ and $O$ are used for the generation of the initial fuzzy rules (Algorithm \ref{SCEFIS_TRAIN}). The initial fuzzy system is built through the creation of a set of rules using the Takagi-Sugeno approach to describe the in- and output matrices. Based on $N_L$ different features from the input and one optimal parameter as the output, a set of rules is generated whereby the features are in the antecedent part and the optimal parameters are in the consequent part of the rules.

\begin{algorithm}[tb]
\caption{Offline and Training Phases}
\begin{algorithmic}[1]
\label{SCEFIS_TRAIN}
\STATE \textbf{------------ Offline phase ------------}
\STATE Determine the parent algorithm(s) and their parameters $p_1,p_2,\cdots,p_k$.
\STATE Read the gold standard images $G_1, G_2, \cdots , G_n$.
\STATE Via exhaustive search or trial-and-error comparisons with gold standard images, determine the best segments $S_1, S_2, \cdots , S_n$ and the best parameters $p_1^*,p_2^*,\cdots,p_k^*$ that generate the best segments and store them in matrix $T$.
\STATE \textbf{------------ Training phase ------------}
\STATE Determine the available training images $I_1, I_2, \cdots , I_{N_R}$.
\STATE \textcolor{black}{Create two empty matrices $M$ for input and $O$ for output.}
\FOR  {all $N_R$ images}

		\STATE Fill matrix $F_T$ with rows from matrix $F^*$ that belong to the training image $I_i$
		($F_T = F^* (I_i) $).
		\STATE Fill matrix $T_R$ with rows from matrix $T$ that belong to the training image $I_i$\\
		($T_R = T(I_i) $).
		\IF {i=1}
		\STATE Append $F_R$ to $M$, and $T_R$ to $O$.
			\ELSE		
	 \STATE Pruning step: Discard rows from $F_R$ and $T_R$ that are similar to rows in $M$ and $O$, respectively.
	 \STATE Append the updated matrices $F_R$ and $T_R$ to $M$ and $O$ respectively.
		\ENDIF
	\ENDFOR

\STATE Generate fuzzy rules $R_{F_1},R_{F_2},\cdots$ from the input matrix $M$ and the output matrix $O$ (e.g., using clustering).

\end{algorithmic}
 \end{algorithm}
 
\subsection{Online and Evolving Phase}
\label{SCEFISTEst}

The evolving process is performed in order to increase the capabilities of the proposed system. For each test image, a matrix $F_S$ is filled with the rows from $F^*$ that belong to the test image (Algorithm \ref{SCEFIS_Use}). Fuzzy inference using $F_S$ is applied, and a parameter vector $T_O$ is returned (size $1 \times 8$) and the final output parameter $T^*$ is calculated (Algorithm \ref{SCEFIS_Use}). The resulting parameter is used for the segmentation of the image (Algorithm \ref{SCEFIS_Use}), and the resulting segment is stored and then displayed to the user for review and eventual correction (Algorithm \ref{SCEFIS_Use}). The best parameter for the current image is then calculated based on the user-corrected segment and is stored in $T_B$ (Algorithm \ref{SCEFIS_Use}). A pruning procedure is performed on $F_S$ and $T_B$ as described in \cite{EFIS}, with the exception that the Euclidean distance thresholds are, in contrast to EFIS, different for different techniques.  After pruning, revised versions of $F_S$ and $T_B$ are appended to $M$ and $O$ (Algorithm \ref{SCEFIS_Use}). In the final step, the current fuzzy inference system, i.e., its rule base, is regenerated using the updated matrices $M$ and $O$ (Algorithm \ref{SCEFIS_Use}), and the process is repeated as long as new images are available.

\begin{algorithm}[tb]
\caption{Online/Evolving Phase }
\begin{algorithmic}[1]
\label{SCEFIS_Use}
\STATE Load the fuzzy rules $R_{F_i}$ and the matrices $M$, $O$, and $F^*$.
\STATE Load the test images $I_1, I_2, \cdots , I_{N_E}$.
\FOR  { all $N_E$ images}
		\STATE Fill matrix $F_S$ with the rows from matrix $F^*$ that belong to the test image $I_i$ ($F_S = F^* (I_i) $).
\STATE Perform fuzzy inference to generate output:\\ $T_O =$ FUZZY-INFERENCE($R_{F_1},R_{F_2},\cdots$).
	
\STATE Generate a single output $T^*$ from $T_O$ using the mean of ${T_O}$ ($\mu_{T_O}$), the median of ${T_O}$ ($M_{T_O}$), the fuzzy membership ($m_{T_O}$) of the standard deviation of $T_O$ ($\sigma_{T_O}$) using a Z-shaped function ($zmf$) \\ $m_{T_O} =  zmf(\sigma_{T_O},[(\mu_{T_O}*0.10) \ \   (\mu_{T_O}*0.20)])$, and \\ $T^* = m_{T_O} * \mu_{T_O} + (1- m_{T_O}) * M_{T_O}$.
\STATE Apply the parameters to segment $I_i$.
\STATE Display segment $S$ and wait for user feedback (user generates a gold standard image $G$ by editing $S$)
\STATE \textbf{--------- *Rule Evolution - Invisible to User* ---------}
\STATE Determine the best output vector $p_1^*,p_2^*,\cdots,p_k^*$ (via comparison of $S$ with $G$) and store it in $T_B$.
 \STATE Pruning -- Discard rows from $F_S$ and $T_B$ that are similar to rows in $M$ and $O$, respectively.
	 \STATE Append the matrices $F_S$ and $T_B$ to $M$ and $O$, respectively.
	\STATE Generate fuzzy rules $R_{F_i}$ from the updated matrices $M$ and $O$ (e.g., using clustering).
	\ENDFOR
\end{algorithmic}
 \end{algorithm}

\section{Experiments and Results}
\label{expResults}
This section describes the experiments conducted in order to test the proposed self-configuring EFIS (SC-EFIS). To build the initial fuzzy system, for each training set, a set of randomly selected images from the data set were used for the extraction of the features along with the optimum parameters as output. This initial fuzzy system was then used to test the proposed method using the remaining images. The initial fuzzy system evolves as long as new (unseen) images are fed into the system and as long as the segmentation results produced by the algorithms are corrected by an expert user in order to generate optimal parameter values. This process drives the evolution of the fuzzy rules for segmentation. We used a 10-fold leave-$n$-patients-out cross-validation for the experimentation. The results of ten different trials for each segmentation technique and for each parent algorithm are presented in order to validate the performance of SC-EFIS. The number of rules was monitored during the evolution process in order to acquire empirical knowledge about the convergence of the evolving process. The experimental results using an image dataset for global thresholding are presented. 

\subsection{Image Data}
\label{Images}
\begin{figure*}[htb]
\center
\includegraphics[width=0.5in,height=0.5in]{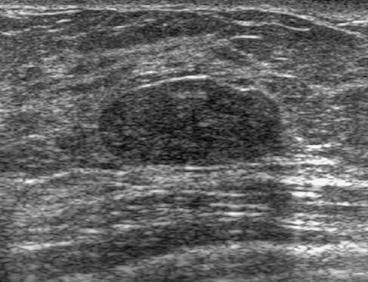}
\includegraphics[width=0.5in,height=0.5in]{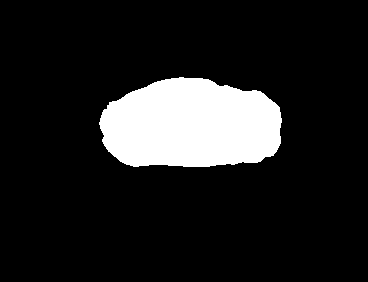}
\includegraphics[width=0.5in,height=0.5in]{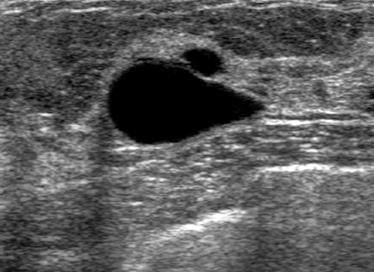}
\includegraphics[width=0.5in,height=0.5in]{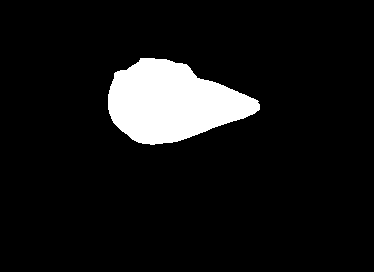}
\includegraphics[width=0.5in,height=0.5in]{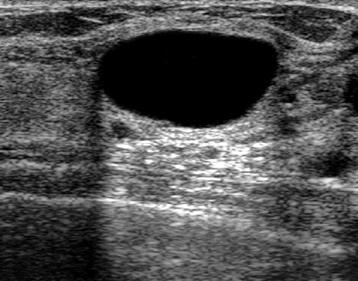}
\includegraphics[width=0.5in,height=0.5in]{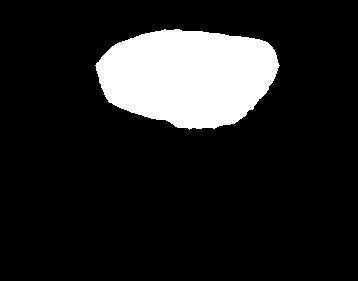}
\includegraphics[width=0.5in,height=0.5in]{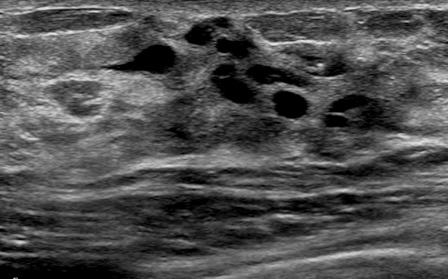}
\includegraphics[width=0.5in,height=0.5in]{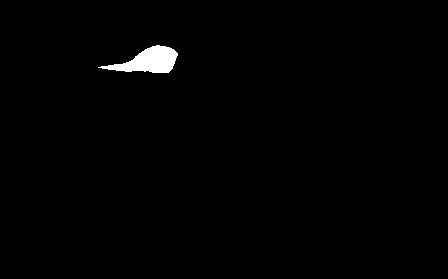}
\includegraphics[width=0.5in,height=0.5in]{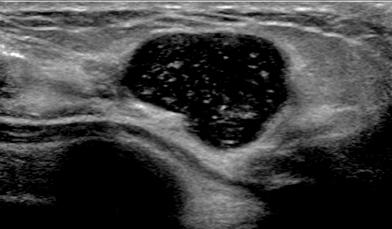}
\includegraphics[width=0.5in,height=0.5in]{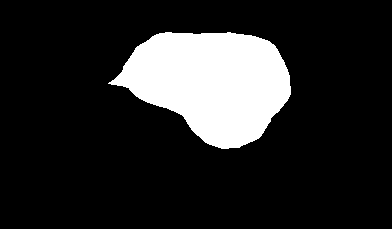} \\ \vspace{0.05in}
\includegraphics[width=0.5in,height=0.5in]{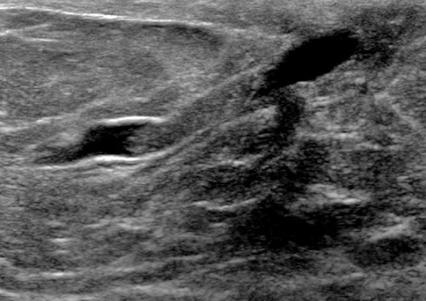}
\includegraphics[width=0.5in,height=0.5in]{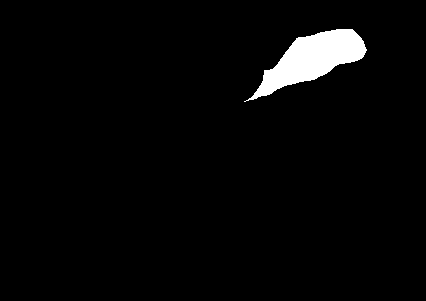}
\includegraphics[width=0.5in,height=0.5in]{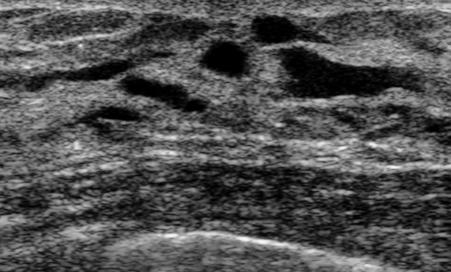}
\includegraphics[width=0.5in,height=0.5in]{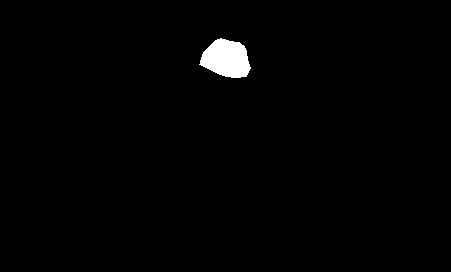}
\includegraphics[width=0.5in,height=0.5in]{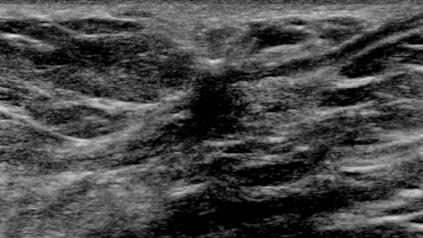}
\includegraphics[width=0.5in,height=0.5in]{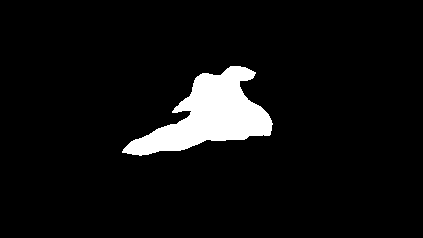}
\includegraphics[width=0.5in,height=0.5in]{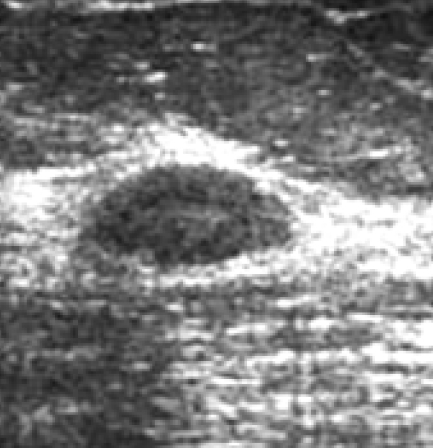}
\includegraphics[width=0.5in,height=0.5in]{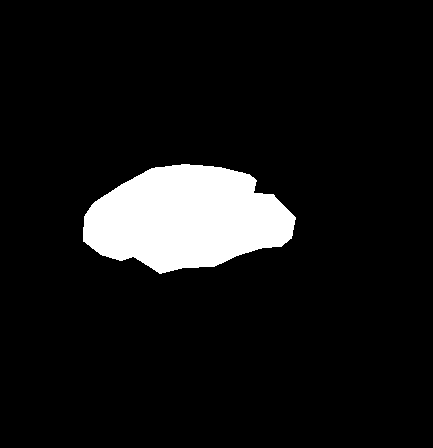}
\includegraphics[width=0.5in,height=0.5in]{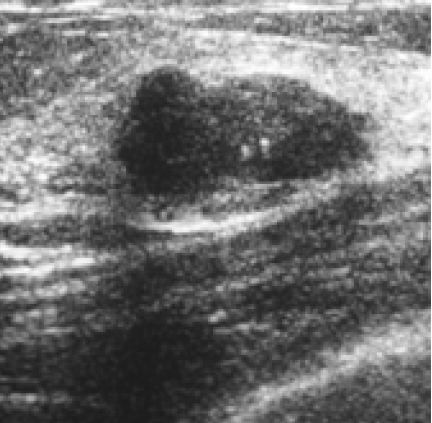}
\includegraphics[width=0.5in,height=0.5in]{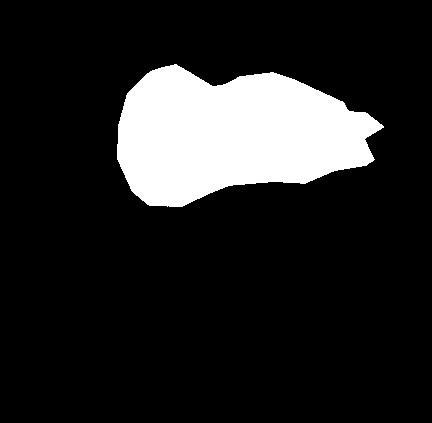} \\ \vspace{0.05in}
\includegraphics[width=0.5in,height=0.5in]{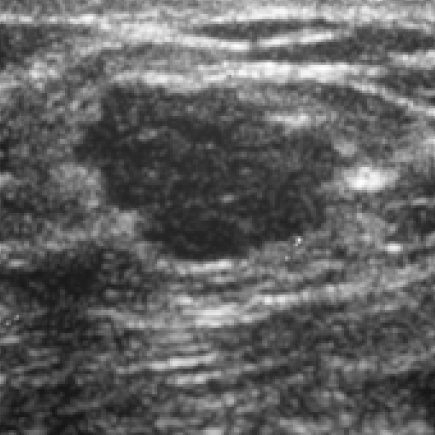}
\includegraphics[width=0.5in,height=0.5in]{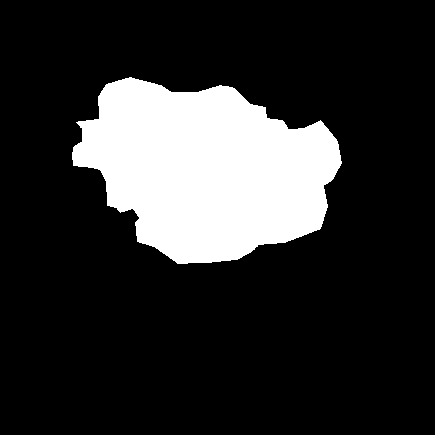}
\includegraphics[width=0.5in,height=0.5in]{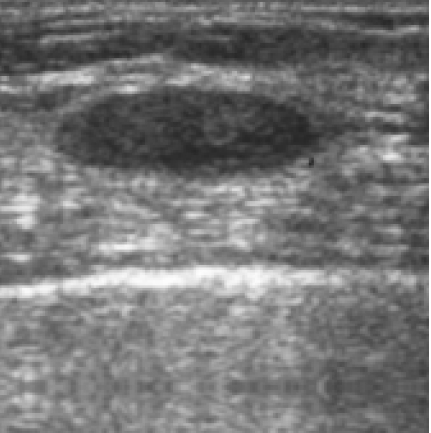}
\includegraphics[width=0.5in,height=0.5in]{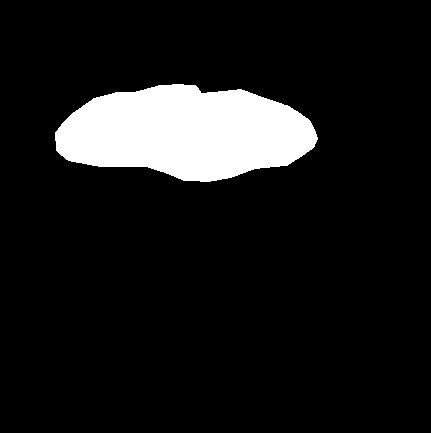}
\includegraphics[width=0.5in,height=0.5in]{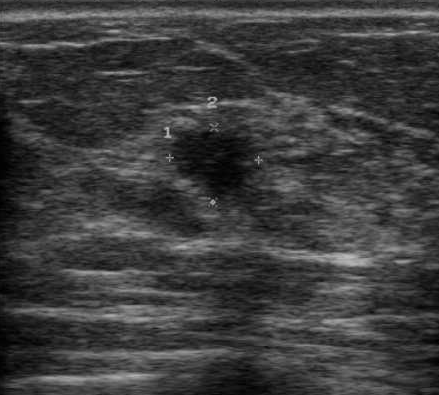}
\includegraphics[width=0.5in,height=0.5in]{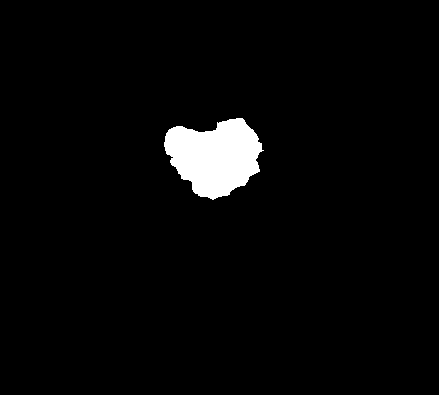}
\includegraphics[width=0.5in,height=0.5in]{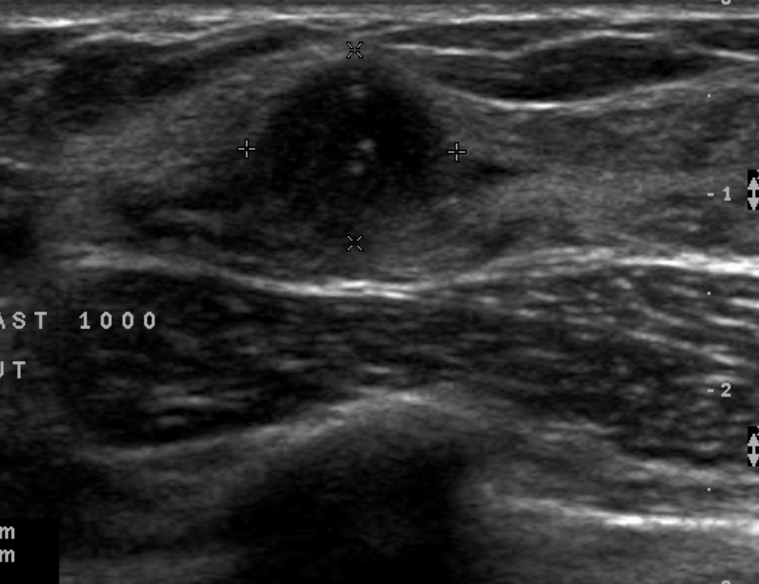}
\includegraphics[width=0.5in,height=0.5in]{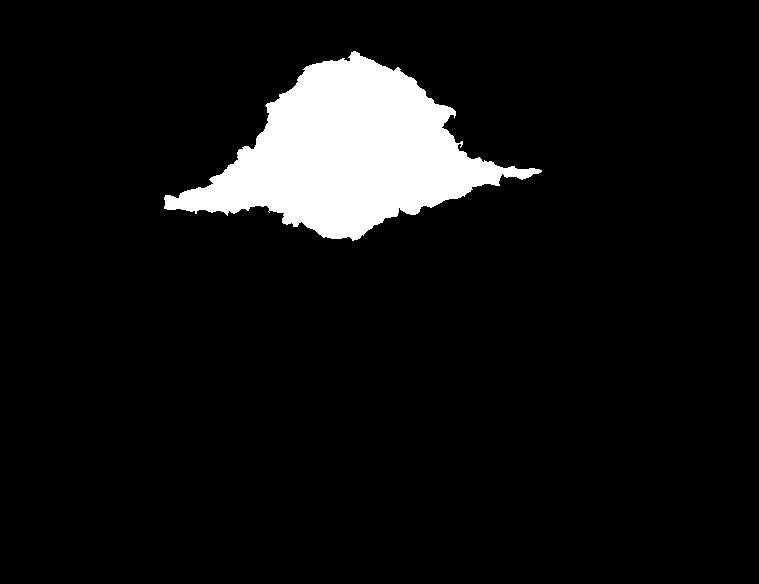}
\includegraphics[width=0.5in,height=0.5in]{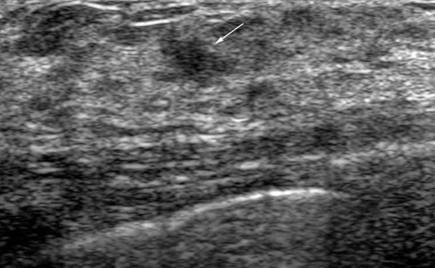}
\includegraphics[width=0.5in,height=0.5in]{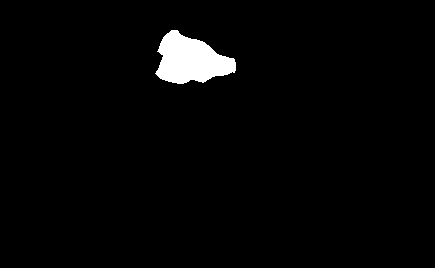} \\ \vspace{0.05in}
\includegraphics[width=0.5in,height=0.5in]{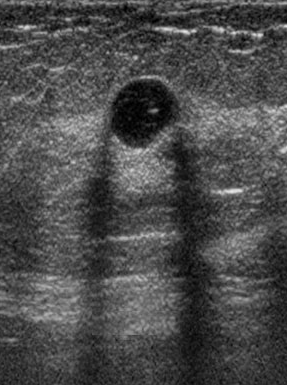}
\includegraphics[width=0.5in,height=0.5in]{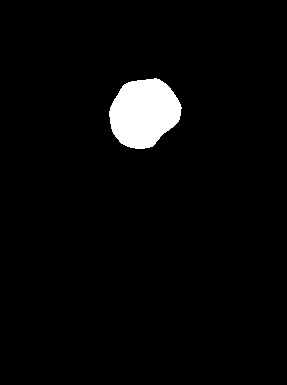}
\includegraphics[width=0.5in,height=0.5in]{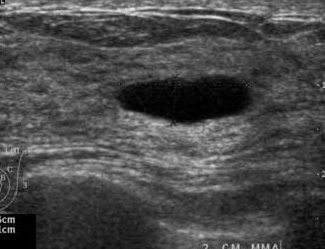}
\includegraphics[width=0.5in,height=0.5in]{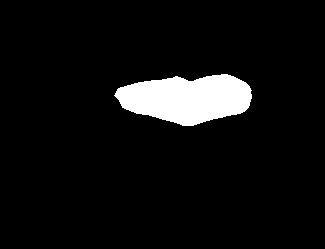}
\includegraphics[width=0.5in,height=0.5in]{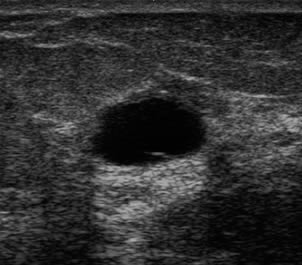}
\includegraphics[width=0.5in,height=0.5in]{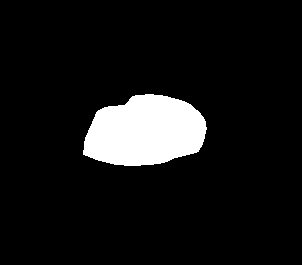}
\includegraphics[width=0.5in,height=0.5in]{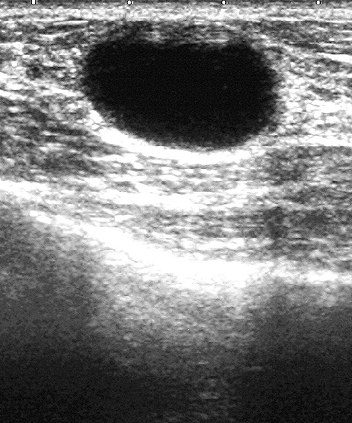}
\includegraphics[width=0.5in,height=0.5in]{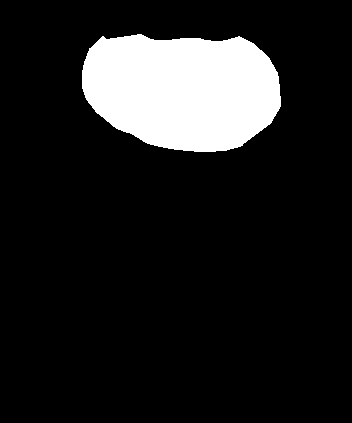}
\includegraphics[width=0.5in,height=0.5in]{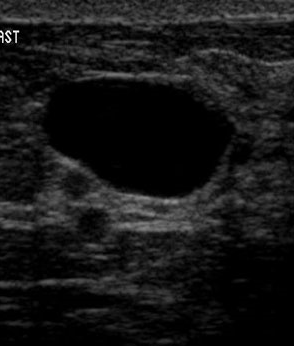}
\includegraphics[width=0.5in,height=0.5in]{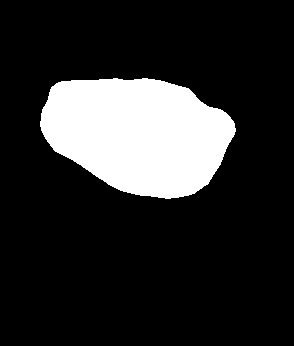} \\ \vspace{0.05in}
\includegraphics[width=0.5in,height=0.5in]{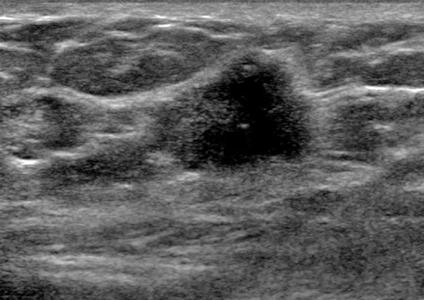}
\includegraphics[width=0.5in,height=0.5in]{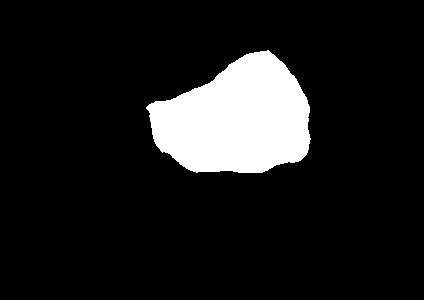}
\includegraphics[width=0.5in,height=0.5in]{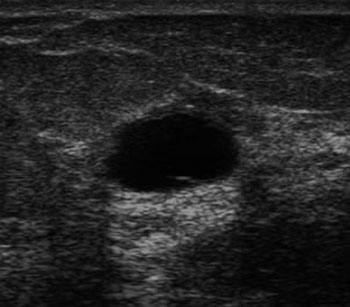}
\includegraphics[width=0.5in,height=0.5in]{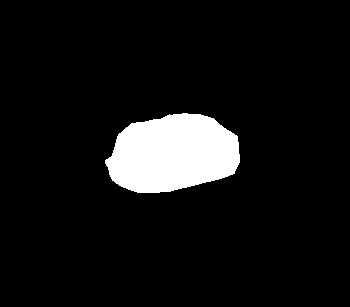}
\includegraphics[width=0.5in,height=0.5in]{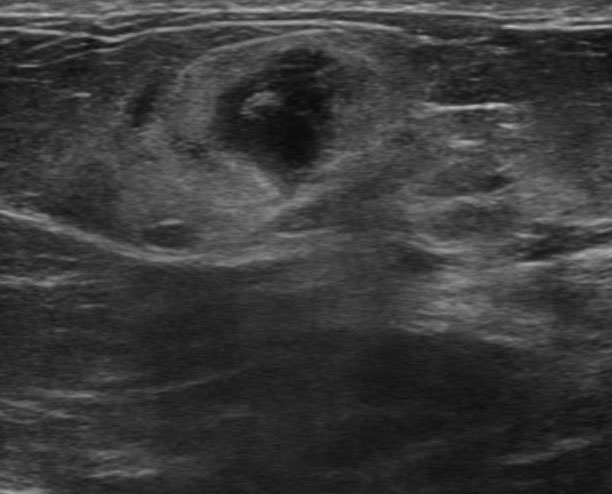}
\includegraphics[width=0.5in,height=0.5in]{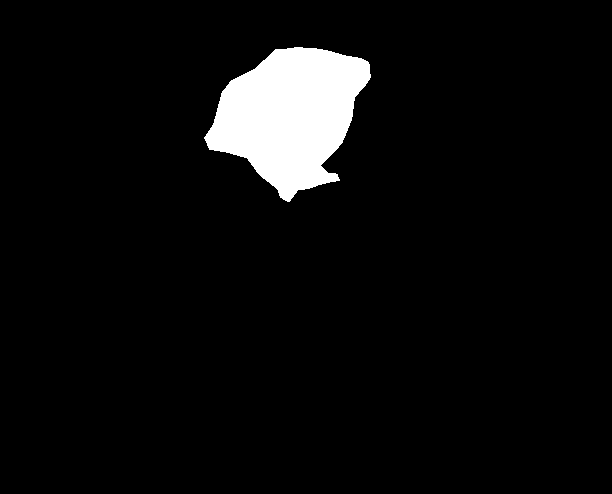}
\includegraphics[width=0.5in,height=0.5in]{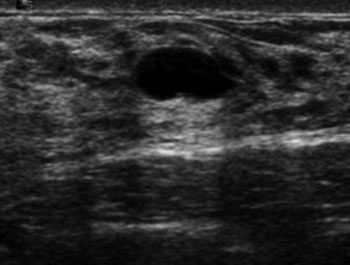}
\includegraphics[width=0.5in,height=0.5in]{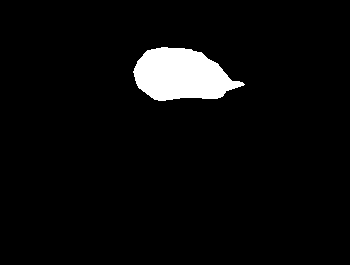}
\includegraphics[width=0.5in,height=0.5in]{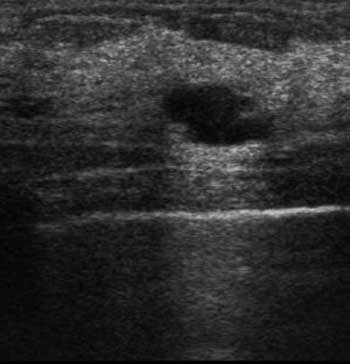}
\includegraphics[width=0.5in,height=0.5in]{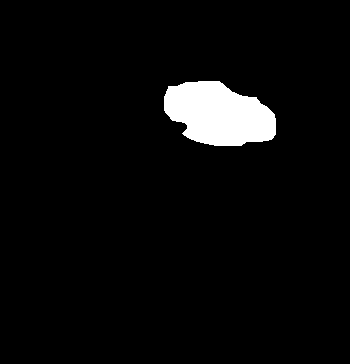} \\ \vspace{0.05in}
\includegraphics[width=0.5in,height=0.5in]{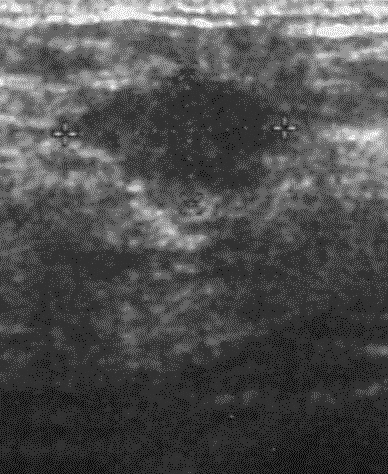}
\includegraphics[width=0.5in,height=0.5in]{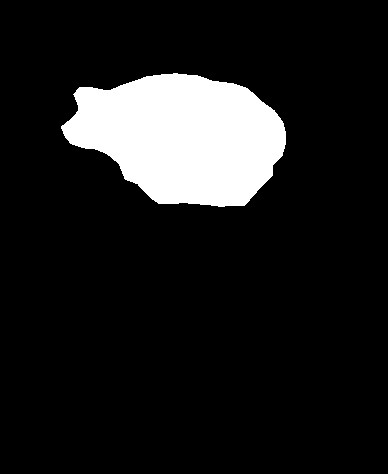}
\includegraphics[width=0.5in,height=0.5in]{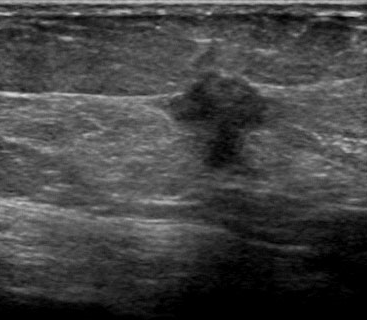}
\includegraphics[width=0.5in,height=0.5in]{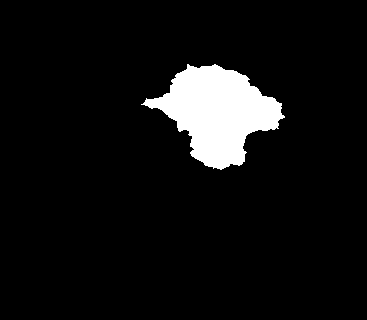}
\includegraphics[width=0.5in,height=0.5in]{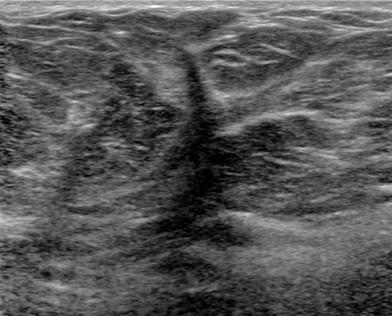}
\includegraphics[width=0.5in,height=0.5in]{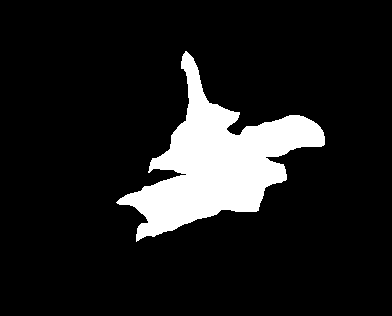}
\includegraphics[width=0.5in,height=0.5in]{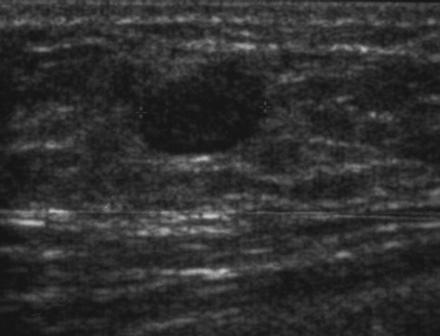}
\includegraphics[width=0.5in,height=0.5in]{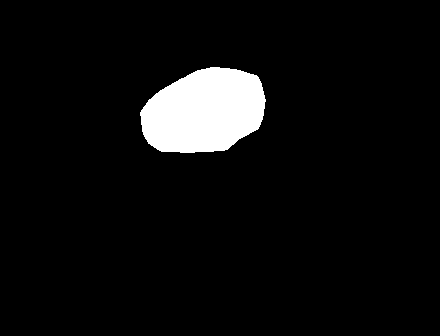}
\includegraphics[width=0.5in,height=0.5in]{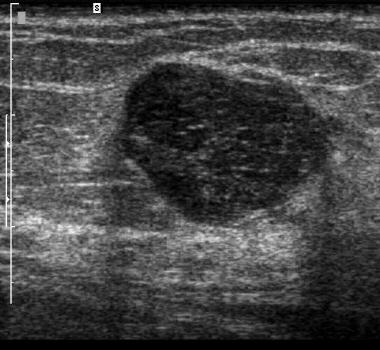}
\includegraphics[width=0.5in,height=0.5in]{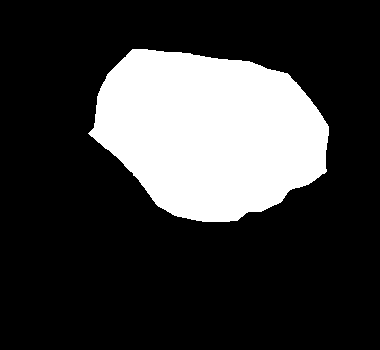} \\ \vspace{0.05in}
\includegraphics[width=0.5in,height=0.5in]{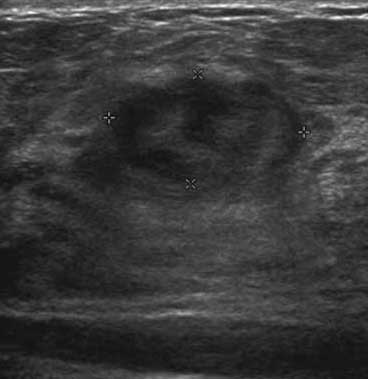}
\includegraphics[width=0.5in,height=0.5in]{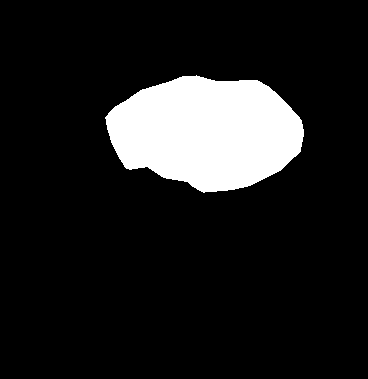}
\includegraphics[width=0.5in,height=0.5in]{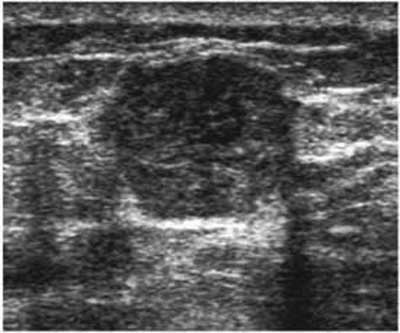}
\includegraphics[width=0.5in,height=0.5in]{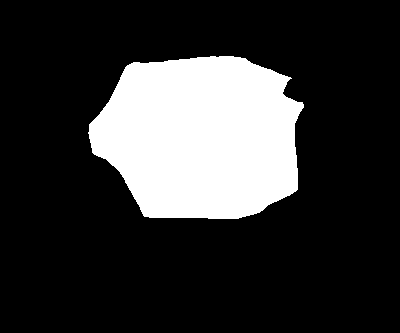}
\includegraphics[width=0.5in,height=0.5in]{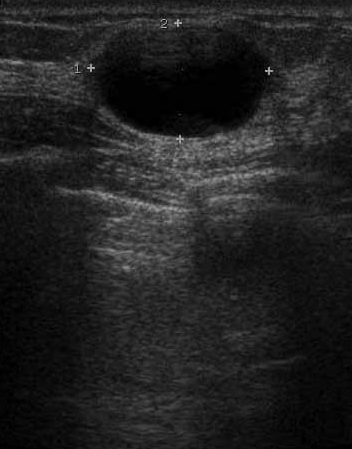}
\includegraphics[width=0.5in,height=0.5in]{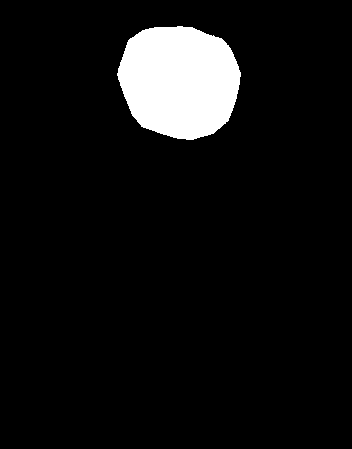}
\includegraphics[width=0.5in,height=0.5in]{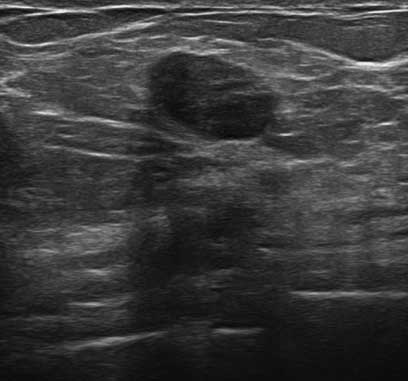}
\includegraphics[width=0.5in,height=0.5in]{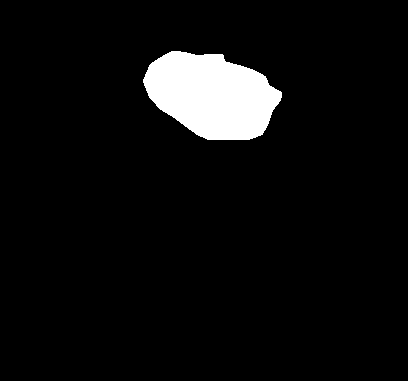}
\includegraphics[width=0.5in,height=0.5in]{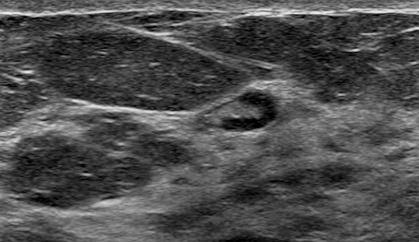}
\includegraphics[width=0.5in,height=0.5in]{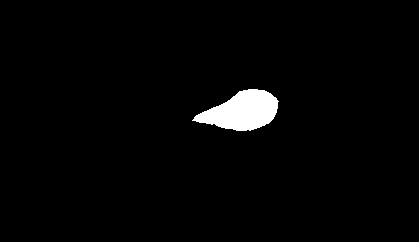}\\
\caption{Breast ultrasound scans used in our experiments. All images were segmented by an image-processing expert.  }
\label{allImages}
\end{figure*}

The target dataset was developed from \textbf{35 breast ultrasound scans} that were segmented by an image-processing expert with extensive experience in breast lesion segmentation (the second author). The images, collected from the Web, are of different dimensions, ranging from $230\times 390$ to $580\times 760$ pixels (Figure \ref{allImages}, images resized for sake of illustration). These are the same images used to introduce EFIS originally \cite{EFIS}.

Ultrasound images are generally difficult to segment, primarily due to the presence of speckle noise and low level of local contrast. It should be noted that the segmentation of ultrasound actually does require a complete processing chain, (including proper preprocessing and post-processing steps). However, the purpose of using these images was solely to demonstrate that the accuracy of the segmentation can be increased with the application of SC-EFIS.

\subsection{Evaluation Measures}
\label{Measure}
Considering two segments $S$ (generated by an algorithm) and $G$ (the gold standard image manually created by an expert), we calculate the average of the Jaccard index $J$ (area overlap) \cite{Tan2005}:
\begin{equation}
\label{jaccardsim}
J(S,G) = \frac{|S\cap G|}{|S\cup G|},
\end{equation}
and its standard deviation $\sigma_J$. As well, the $95\%$ confidence interval (CI) of the Jaccard index $CI_J$ is calculated . Finally, we performed t-tests to validate the null hypothesis for comparing the results of a parent algorithm and its evolved version in order to establish whether any potential increase in accuracy is statistically significant. Ground-truth images $G$ were created so that the objects of interest (i.e., lesions and tumours) could be labeled as white (1) and the background as black (0). All thresholding techniques were used consistently to label object pixels in this way as this was done in EFIS. We also calculate ``Maximum Achievable Accuracy'' (MAA) determined via exhaustive search and through comparison with gold standard images. MAA of each i hence the maximum accuracy that any global thresholding method can achieve.

\subsection{Results}
 \label{SCEFIS}

To compare with EFIS, the SC-EFIS results are calculated for global thresholding. The results are discussed with respect to rule evolution, and accuracy verification using the Jaccard results. As well, we compare the results of EFIS and SC-EFIS with other methods such as local thresholding by Niblick \cite{nib}, the Huang/Wang method for fuzzy thresholding \cite{Huang1995}, the Kittler algorithm \cite{Kittler1986}, the interval-based fuzzy thresholding according to Tizhoosh \cite{Tizhoosh2005image,Tizhoosh2008inbook} and the Otsu method. It should be noted that comparing with the Niblack method is actually not a correct comparison because it is a local method whereas all other methods operate globally, so it is expected that Niblack should generally outperform global methods.  

\begin{figure}[t]
\includegraphics[width=0.90\columnwidth]{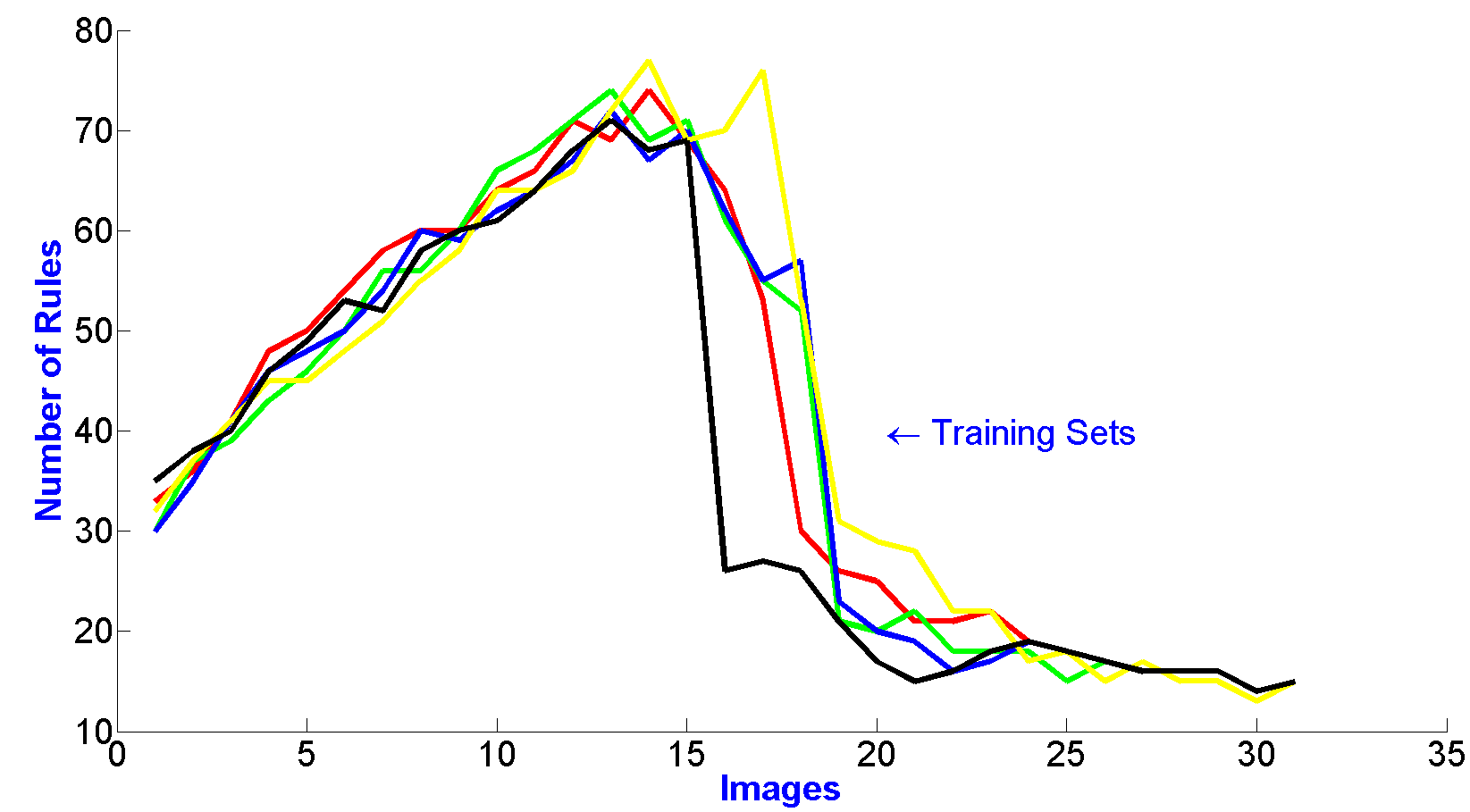}
\caption{Rule evolution for SC-EFIS for thresholding: The number of rules increases first as more images are processed but then drops and seems to converge toward a lower number of rules. Each curve shows the number of rules for a separate trial/run.}
\label{Ruleevolve}
\end{figure}

 \textbf{Rule Evolution} -- Fig. \ref{Ruleevolve} indicates the change in the number of rules during the evolving of the thresholding (THR) process. The initial number of rules increases with any incoming image and then begins to decrease as additional images become available. \textbf{Accuracy Verification} -- Ten different trials/runs are presented for each method. Each run is an independent experiment involving different training and testing images.
Table \ref{SCCOMPresults} enables a comparison of EFIS and SC-EFIS results for global thresholding with different global and local thresholding techniques. 
It is clear that, in the three experiments, EFIS and SC-EFIS provide outcomes that are more accurate than those produced with the non-evolutionary thresholding techniques. As well, if the results for SC-EFIS are sightly lower than EFIS (e.g. in the 2nd run, Table \ref{SCCOMPresults}), it is statistically significant. Besides one should be keep in mind that SC-EFIS does not have EFIS limitations.


\begin{table}[t]
\center
\caption{Comparison of EFIS, SC-EFIS, and 4 other global thresholding technique as well as one local thresholding method (\cite{Tizhoosh2005image,Tizhoosh2008inbook, nib, Kittler1986,Huang1995}): Average and standard deviation of the Jaccard index $J \pm \sigma_J$ and 95\% confidence interval $CI_J$. }
\begin{tabular}{|c|l|c|c|c|}

\hline
Run	& Method	& $J \pm \sigma_J$ & $CI_J$  \\ \hline
	& \cellcolor[gray]{0.8} MAA 		&  \cellcolor[gray]{.8} 79\%$\pm$12\% & \cellcolor[gray]{.8} [75\%	84\%]  \\
	& \textbf{EFIS-THR}	&  \textbf{62\%$\pm$25\%} & \textbf{[53\%	71\%]} \\
	& \textbf{SC-EFIS-THR}	&  \textbf{63\%$\pm$23\%} & \textbf{[54\%	72\%]} \\
	& Niblack	(local)		&  56\%$\pm$24\% & [47\%	65\%] \\
1	& Huang			&  45\%$\pm$27\% & [35\%	55\%] \\
	& Kittler			&  39\%$\pm$32\% & [27\%	51\%] \\
	& Tizhoosh		&  35\%$\pm$32\% & [23\%	47\%] \\
	& Otsu     			&  28\%$\pm$25\% & [18\%	37\%] \\ \hline
	& \cellcolor[gray]{0.8}  MAA 		&  \cellcolor[gray]{0.8}  79\%$\pm$11\% & \cellcolor[gray]{0.8}  [75\%	83\%]  \\
	& \textbf{EFIS-THR}	&  \textbf{60\%$\pm$24\%} & \textbf{[51\%	69\%]} \\
		& \textbf{SC-EFIS-THR}	&  \textbf{59\%$\pm$26\%} & \textbf{[49\%	69\%]} \\
	& Niblack	(local)		&  57\%$\pm$25\% & [48\%	66\%] \\
2	& Huang			&  44\%$\pm$29\% & [34\%	55\%] \\
	& Kittler			&  41\%$\pm$31\% & [29\%	52\%] \\
	& Tizhoosh		&  38\%$\pm$32\% & [26\%	50\%] \\
	& Otsu     			&  29\%$\pm$25\% & [19\%	38\%] \\ \hline
	& \cellcolor[gray]{0.8}  MAA 		&  \cellcolor[gray]{0.8}  79\%$\pm$12\% & \cellcolor[gray]{0.8}  [74\%	83\%]  \\
	& \textbf{EFIS-THR}	&  \textbf{63\%$\pm$23\%} & \textbf{[54\%	71\%]} \\
			& \textbf{SC-EFIS-THR}	& \textbf{65\%$\pm$21\%} & \textbf{[57\%	73\%]} \\
	& Niblack	(local)		&  59\%$\pm$24\% & [49\%	68\%] \\
3	& Huang			& 46\%$\pm$27\% & [35\%	56\%] \\
	& Kittler			&  41\%$\pm$33\% & [29\%	53\%] \\
	& Tizhoosh		&  35\%$\pm$33\% & [23\%	48\%] \\
	& Otsu     			&  28\%$\pm$23\% & [20\%	37\%] \\
\hline
\end{tabular}
\label{SCCOMPresults}
\end{table}
\section{Conclusions}
\label{CON}
Evolving fuzzy image segmentation (EFIS) has been recently proposed to provide evolving and user-oriented adjustment. EFIS is a generic segmentation scheme that relies on user feedback in order to improve the quality of segmentation. Its evolving nature makes this approach attractive for applications that incorporate high-quality user feedback, such as in medical image analysis. 

 However, EFIS entails some limitations, such as parameters that must be selected prior to the running of the algorithm and the lack of an automated feature selection component. These drawbacks restrict the use of EFIS to specific categories of images. 
 An improved version of EFIS, called self-configuring EFIS (SC-EFIS) was proposed in this paper. SC-EFIS is a generic image segmentation scheme that does not require setting of most parameters, such as number of features or detecting a region of interest. SC-EFIS operates with the data available and extracts major parameters necessary for its operation from those data. A comparison of the SC-EFIS results with those obtained with EFIS demonstrates the comparable accuracy of both schemes with SC-EFIS offering a much higher level of automation.

\bibliographystyle{ieeetr}
\bibliography{ref4}
\end{document}